\documentclass[runningheads]{llncs}

\usepackage{eccv}

\usepackage{eccvabbrv}

\usepackage{graphicx}
\usepackage{booktabs}
\definecolor{darkgreen}{RGB}{0,102,0}
\usepackage{colortbl}
\usepackage{multirow}

\usepackage[accsupp]{axessibility}  

\usepackage{hyperref}

\usepackage{orcidlink}

\DeclareMathOperator*{\argmax}{argmax}

\usepackage[dvipsnames]{xcolor}

\usepackage{algorithm}
\usepackage{algpseudocode}
\usepackage{multirow}
\usepackage{arydshln}

\definecolor{Gray}{gray}{0.85}

\usepackage{graphicx}
\usepackage{amsmath}
\usepackage{amssymb}
\usepackage{booktabs}
\definecolor{darkgreen}{RGB}{0,102,0}

\def\ie{\emph{i.e}\onedot}

\usepackage{amsmath}
\usepackage{bbm}
\usepackage{multirow}

\begin{document}

\title{Finding Meaning in Points: Weakly Supervised Semantic Segmentation for Event Cameras} 

\titlerunning{Finding Meaning in Points: WSSS for Event Cameras}

\newcommand\CoAuthorMark{\footnotemark[\arabic{footnote}]}

\author{Hoonhee Cho\orcidlink{0000-0003-0896-6793}\thanks{Equal contribution.}, Sung-Hoon Yoon\orcidlink{0000-0001-5851-2031}\protect\CoAuthorMark, Hyeokjun Kweon\orcidlink{0000-0003-4442-5513}\protect\CoAuthorMark, \\
and Kuk-Jin Yoon\orcidlink{0000-0002-1634-2756}}

\authorrunning{Cho et al.}

\institute{Visual Intelligence Lab., KAIST \\
\email{\{gnsgnsgml, yoon307, 0327june, kjyoon\}@kaist.ac.kr}
}

\maketitle
\begin{abstract}
  Event cameras excel in capturing high-contrast scenes and dynamic objects, offering a significant advantage over traditional frame-based cameras. Despite active research into leveraging event cameras for semantic segmentation, generating pixel-wise dense semantic maps for such challenging scenarios remains labor-intensive. As a remedy, we present EV-WSSS: a novel weakly supervised approach for event-based semantic segmentation that utilizes sparse point annotations. To fully leverage the temporal characteristics of event data, the proposed framework performs asymmetric dual-student learning between 1) the original forward event data and 2) the longer reversed event data, which contain complementary information from the past and the future, respectively. Besides, to mitigate the challenges posed by sparse supervision, we propose feature-level contrastive learning based on class-wise prototypes, carefully aggregated at both spatial region and sample levels. Additionally, we further excavate the potential of our dual-student learning model by exchanging prototypes between the two learning paths, thereby harnessing their complementary strengths. With extensive experiments on various datasets, including DSEC Night-Point with sparse point annotations newly provided by this paper, the proposed method achieves substantial segmentation results even without relying on pixel-level dense ground truths. The code and dataset are available at \url{https://github.com/Chohoonhee/EV-WSSS}.
  \keywords{Event camera \and Weakly supervised semantic segmentation}
\end{abstract}

\section{Introduction}
\label{sec:intro}

Event cameras have recently gained popularity in computer vision and robotics due to their unique features, such as exceptional dynamic range~\cite{Rebecq2019HighSA, MostafaviIsfahani2018EventBasedHD}, ultra-fast response times~\cite{Gehrig2022AreHE, Falanga2020DynamicOA}, and robustness against motion blur~\cite{Huang2023EventbasedSL, cho2023non, kim2024frequency}. 
Their application in automotive technology is on the rise, offering solutions for difficult scenarios like low-light environments~\cite{cho2022selection, cho2024tta, liang2023coherent}, transitioning from dark tunnels into bright sunlight~\cite{Gehrig2021DSECAS}, or rapid motion in drones~\cite{Falanga2020DynamicOA, zhu2018multivehicle, cho2022event}.
Therefore, event-based semantic segmentation is expected to significantly improve system reliability and safety, capitalizing on the characteristics of resilience to varying lighting conditions and their quick response times.

However, event-based semantic segmentation is still in its infancy. 
In our view, the biggest hurdle in achieving substantial event-based semantic segmentation lies in the scarcity of labels.
This stems from the difficulty of obtaining dense ground truth (GT) semantic segmentation maps corresponding to event data.
Although several autonomous driving event-based semantic segmentation datasets~\cite{Alonso2018EVSegNetSS,Gehrig2021CombiningEA,Sun2022ESSLE,Binas2017DDD17ED} have recently emerged, the practical application remains a distant goal. 
A common assumption in these datasets is the availability of images aligned with event data, during annotation process.
However, this expectation often does not hold in real-world scenarios~\cite{Sun2022ESSLE, Wang2021EvDistillAE}, stemming from either the absence of image data or the complexities involved in thoroughly annotating images obtained in edge cases, such as those involving low-light situations or dynamic objects with motion blur.
Further, even if we can access images, acquiring dense semantic segmentation GT is still expensive and labor-intensive.

\begin{figure*}[t]
    \centering\includegraphics[width=1.0\linewidth]{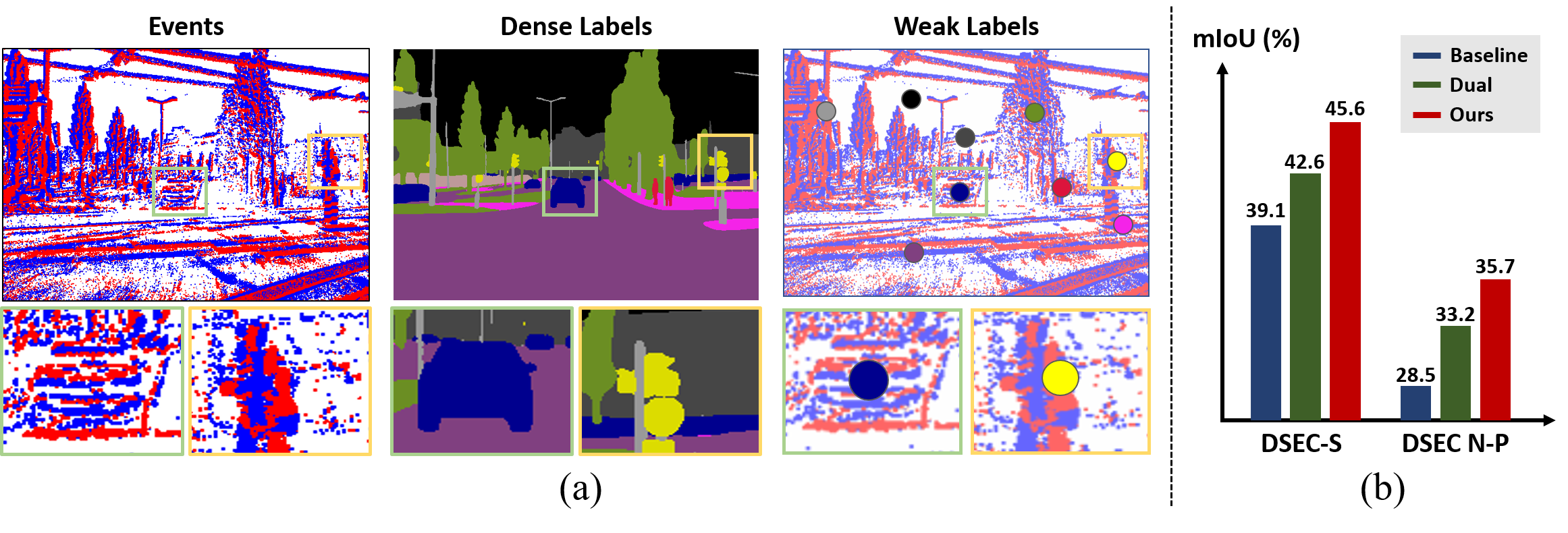}
    \caption{(a) Motivation of the event-based weakly supervised semantic segmentation. (b) Performance comparisons between our baseline, baseline with dual-student learning, and our final model on DSEC-Semantic~\cite{Sun2022ESSLE} and DSEC Night-Point datasets.}
    \label{fig:teaser}
\end{figure*}

Under the practical event-only setting, distinguishing boundaries between objects or accurately segmenting tiny objects on asynchronous event data is extremely challenging. 
Events are primarily triggered at edges where intensity changes frequently, which facilitates boundary detection but can lead to blurry events in over-triggered areas due to excessive events. As shown in Fig.~\ref{fig:teaser}(a), while event data can sufficiently capture the semantic information of objects, obtaining the dense GT semantic segmentation map is challenging due to difficulties in defining precise boundaries.
The images in the lower row of Fig.~\ref{fig:teaser}(a) depict areas where events are insufficiently triggered or excessively fired, leading to saturation. Therefore, labeling dense labels is a significant challenge from a human perspective. Conversely, identifying an object's class is relatively straightforward for humans, and it is much easier to convey information at the class level. 

From these observations, we introduce event-based weakly supervised semantic segmentation (EV-WSSS) for the first time to enhance the practicality and applicability of event-based semantic segmentation. 
Conventional works on WSSS have been actively conducted in the image domain to reduce the burden of the dense annotation process.
By utilizing easily obtainable weak labels such as class tags~\cite{ahn2018learning,wang2020self,zhang2020reliability,fan2020learning,ahn2019weakly,chang2020weakly,lee2021anti,kweon2021unlocking,zhang2021complementary,li2021pseudo,xie2022c2am,zhou2022regional,chen2022self,yoon2022adversarial,mat,chen2023fpr,peng2023usage,xu2022multi,beco,ocr,kweon2023weakly,yoon2021exploring,kweon2024sam,yoon2024class}, points~\cite{fan2022pointly,cheng2022pointly,tang2022active,liao2023attentionshift, kim2022beyond}, scribbles~\cite{lin2016scribblesup,vernaza2017learning}, and bounding boxes~\cite{dai2015boxsup,khoreva2017simple,papandreou2015weakly,lee2021bbam}, WSSS methods have shown promising results.
Motivated by this, our EV-WSSS employs a point-based labeling strategy termed ``1-class-1-click (1C1C)'', providing a single point annotation per each existing class, as in Fig.~\ref{fig:teaser}.

While using the weak point labels instead of dense labels greatly alleviates the annotation burden in event-based semantic segmentation, it exhibits a new challenge in that both the events and the labels (\textit{i.e.}, point supervision) are sparse. 
Various WSSS works~\cite{ahn2018learning,ahn2019weakly,zhang2020reliability,kweon2021unlocking} in the imagery domain have proposed generating dense pseudo-labels from model predictions to handle the lack of supervision.
However, unlike the image data that provides rich and dense features, the inherently sparse nature of event stream data leads to increased ambiguity in semantic inference during pseudo-label generation.
Therefore, the main challenge of EV-WSSS is to compensate for 1) the potential spatial deficiencies in events and 2) the lack of dense supervision in weak labels, at the same time. 

We tackle this problem using two distinct approaches and further integrate them into one powerful framework, leading to much better feature representation.
Firstly, we devise a dual-student learning scheme, utilizing the temporal properties of events.
Considering that the event data is a continuous stream of information captured across the dynamic scene, it is intuitive that the forward flow from the past and the backward flow from the future would have complementary benefits.
For example, this trait possibly facilitates our segmentation model learning to identify the overall outlines and tiny objects~\cite{nam2022stereo}.
To effectively leverage the potential from the data-centric perspective, we formulate our framework as a mutual learning between forward and backward branches, while the prediction of one branch serves as a pseudo-GT of the other at the logit level.

Secondly, to further compensate for the sparse point supervision, we formulate a feature-level prototype-based contrastive learning.
Specifically, we aggregate the features into class-wise prototypes, which are robust representations for each class, considering both the reliable regional information in a single event data and the variety innated in multiple event data.
Finally, we amalgamate the benefits of the dual-student learning framework and prototype-based contrasting approach via prototype-level distillation.
This approach mainly aims to deliver concentrated semantic information from one branch to the other by projecting the aggregated prototypes, while preserving the distinct latent spaces of each branch which are individually optimized for the temporally different event data.

We experimentally verify the proposed EV-WSSS framework on DDD17~\cite{Alonso2018EVSegNetSS} and DSEC-Semantic~\cite{Sun2022ESSLE}, standard benchmarks for event-based segmentation.
The results show that our dual-student scheme with prototype-based contrastive learning brings remarkable performance gains compared to the baseline, as in Fig.~\ref{fig:teaser}(b).
Further, to clearly demonstrate the benefit of our weakly supervised approach under an event-only setting, we build a novel dataset named DSEC Night-Point. 
This dataset is composed of the event data from DSEC Night-Semantic~\cite{Xia2023CMDACD, Gehrig2021DSECAS}, where each data is annotated with our 1C1C setting.
Experimental results on DSEC Night-Point support that EV-WSSS is a promising solution for event-based segmentation when dense labeling is not available.

\section{Related Works}
\label{sec:related_works}

\subsection{Event-based Semantic Segmentation}

Event-based semantic segmentation aims to interpret scenes by utilizing temporal continuity. DDD17, the first benchmark for event-based semantic segmentation, was introduced in \cite{Alonso2018EVSegNetSS, Binas2017DDD17ED}. 
Developments like HALSIE~\cite{Biswas2022HALSIEHA} and HMNet~\cite{Hamaguchi2023HierarchicalNM} have propelled the progress in event-based semantic segmentation, concentrating on the integration of cross-domain features and the utilization of memory-based event analysis.
DTL~\cite{Wang2021DualTL} achieves performance gains through better feature representation learning by jointly learning with image reconstruction.
On the other side, the use of spiking neural networks~\cite{Cordone2021LearningFE, Fang2020IncorporatingLM,ParedesValls2021SelfSupervisedLO} is gaining traction designed for reduced latency via surrogate gradient techniques~\cite{Che2022DifferentiableHA}.

However, a major hurdle for event-based semantic segmentation is the requirements on dense semantic segmentation labels, which are cost-expensive to obtain.
Existing works have attempted to address this issue in two main directions.
One approach involves generating synthetic events from videos~\cite{Gehrig2019VideoTE} or static images~\cite{Messikommer2021BridgingTG}, utilizing extra labeled datasets containing these videos or images to facilitate training through a labeled synthetic event dataset.
Another approach is similar to unsupervised domain adaptation~\cite{Saporta2022MultiHeadDF, cho2023learning, Gao2022CrossDomainCD,Zou2018UnsupervisedDA,Pan2020UnsupervisedIA,Zhang2019CategoryAU,Wang2020DifferentialTF,Xia2023CMDACD,cho2023label}, where the target domain consists of unlabeled event data, and the source domain comprises labeled image data, facilitating the transfer of knowledge. 
Some of these approaches assume that images paired with the event data exist in the target domain.
On the other hand, ours utilizes event data only, without relying on any image, even in the annotation phase.
Previous studies have proposed various approaches to address the annotation challenge in event semantic segmentation, and we also suggest an approach to reduce the reliance on event labels.
Unlike existing methods, we aim to address the issue with weak labels in the event domain without accessing the source domain.

\subsection{Weakly Supervised Semantic Segmentation}
The annotation process for obtaining pixel-wise ground truth (GT) semantic segmentation maps is notoriously labor-intensive.
Against this background, weakly supervised semantic segmentation has been extensively explored to leverage weak yet inexpensive labels.
Throughout the last decade, the WSSS literature for the imagery domain has explored various weak labels, including image-level ~\cite{ahn2018learning,wang2020self,zhang2020reliability,fan2020learning,ahn2019weakly,chang2020weakly,lee2021anti,kweon2021unlocking,zhang2021complementary,li2021pseudo,xie2022c2am,zhou2022regional,chen2022self,yoon2022adversarial,mat,chen2023fpr,peng2023usage,xu2022multi,beco,ocr}, points~\cite{fan2022pointly,cheng2022pointly,tang2022active,liao2023attentionshift, kim2022beyond}, scribbles~\cite{lin2016scribblesup,vernaza2017learning}, and bounding boxes~\cite{dai2015boxsup,khoreva2017simple,papandreou2015weakly,lee2021bbam}.
As these weak labels do not provide explicit information about the segment, the existing WSSS works have usually focused on learning which pixels should be grouped into one segment. 

One of the most extensively studied settings is using image-level class tags, which exploit the information about which classes exist in each image.
These approaches first train an image-level classifier using the class tags and then obtain Class Activation Maps (CAMs)~\cite{zhou2016learning} from the classifier as an initial seed for semantic segmentation.
However, if every image in the training dataset includes a certain class, \textit{e.g.}, road, a classifier is simply biased to that class rather than to learning the concept correctly.
Recently, a method using the CLIP~\cite{clip} has been proposed~\cite{kim2023weakly}; however, this approach cannot be directly applied to our event-only approach.

Another actively studied setting is sparse point supervision.
This setting assumes that every instance in the image is annotated by a single (or a pre-defined number of) points with class annotation.
Although instance-wise point supervision has been hugely successful in the imagery domain, we find it difficult to identify and annotate every instance of event data, especially for tiny objects.
Therefore, in this paper, we present a 1-class-1-click (1C1C) setting, which assumes at least one pixel is annotated for each class existing in the event data.

\section{Methods}
\subsection{Problem Formulation}

\subsubsection{Weak Labels: 1-class-1-click.}
In the proposed EV-WSSS framework, we present 1-class-1-click (1C1C).
As event data is sparsely distributed in both spatial and temporal axes, directly labeling it is difficult.
Therefore, we visualize the event stream data by stacking it into a frame-like representation for the annotation process, as depicted in Fig.~\ref{fig:teaser}.
In the annotation process, we first request annotators to identify which classes exist in the scene.
Subsequently, for each existing class, the annotators manually click a single pixel of the visualized event data.
The obtained labels are represented as a set of point annotations $\mathbf{t}=\{(x_i,y_i,c_i)\}$. Here, $x_i, y_i$ denotes the spatial coordinates of each point, and $c_i \in \{1,\ldots,C\}$ represents its class, where $C$ is the total number of classes.

\noindent
\textbf{Event Representation and Embedding.}
An event camera asynchronously detects log intensity changes \(\Delta \log(I_{x,y,t})\) for each pixel, triggering events \(e_{x,y,t,p}\) when changes exceed a threshold, thus capturing intensity continuously with microsecond latency. Polarity \(p\) indicates the direction of change, with \(1\) for positive and \(-1\) for negative changes. 
To perform event-based vision tasks at a given time step \(T\), the event stream \(\mathcal{E}_{(T - \tau) \rightarrow T} = \{e_{x_n,y_n,t_n,p_n}\}_{n=0}^{N-1}\), from \(T - \tau\) to \(T\) with \(N\) events over a time period of \(\tau\), is stacked to adapt it for neural networks and converted into the voxel grid~\cite{Zhu2018UnsupervisedEL, Sun2022ESSLE}, $E_{(T - \tau) \rightarrow T}$.
We define the event voxel grid, $E_{(T - \tau) \rightarrow T}$, generated in this manner as the forward voxel grid.
In addition to the forward voxel grid, our framework described in Section~\ref{method:dual} utilizes the future event stream, which are exclusively available during the only training phase.
Specifically, we stack the event stream \(\mathcal{E}_{T \rightarrow (T+\tau')}\) beyond the target time step \(T\) over a period of $\tau'$.
The stacked future events represent not an inference for time $T$ but for $T + \tau'$. 
To handle this, we use event sequence reversal~\cite{Tulyakov2021TimeLE, He2022TimeReplayerUT} to model the real event camera as closely as possible while simultaneously aligning the target time with $T$, just like the forward event stream.
We name the obtained future-and-reversed as the ``backward event stream'', and for simplification, we will denote the forward and backward event voxel grids as \(E^f\) and \(E^b\), respectively.

To effectively utilize the continuous temporal information of events, we employ a recurrent encoder~\cite{Sun2022ESSLE} for both the forward and backward encoders, $\phi^f$ and $\phi^b$. The temporally accumulated features of forward and backward events, $\mathbf{Z}^f$ and $\mathbf{Z}^b$, through the \(\phi^f\) and \(\phi^b\) encoders are obtained as follows, respectively:
\begin{equation}
\mathbf{Z}^f=\phi^f\left(E^f\right) \quad \text { and } \quad \mathbf{Z}^b=\phi^b\left(E^b\right).
\end{equation}

\begin{figure*}[t]
    \centering
    \includegraphics[width=.95\linewidth]{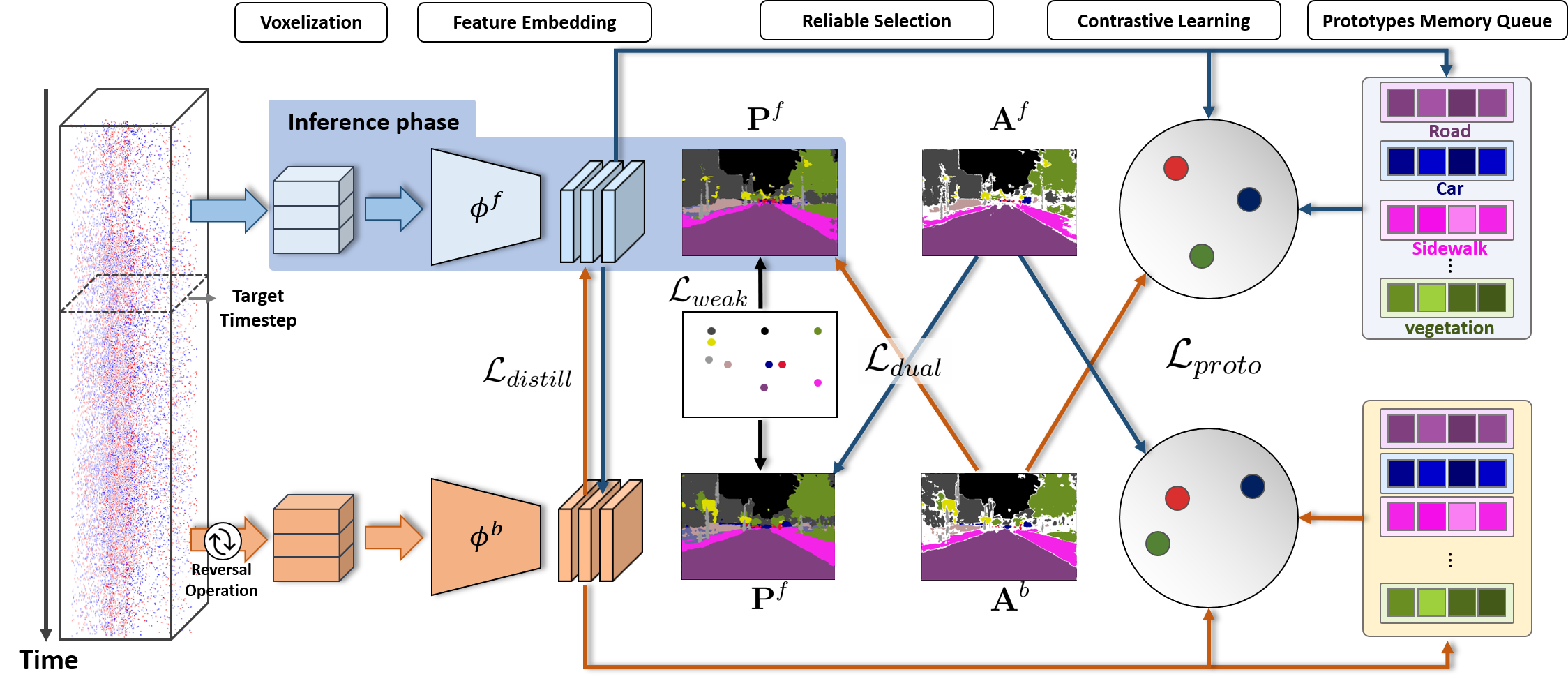}
    \caption{Overview of the proposed EV-WSSS framework. We omit the details about prototype-related components in this figure for better understanding.}
    \label{fig:overall_framework}
\end{figure*}

\subsection{Asymmetric Dual-Student Learning}
\label{method:dual}
Events operate asynchronously based on changes in light intensity, allowing the acquisition of event streams. This asynchronous operation offers the advantage of diverse representations of specific timestamps, achievable by stacking events in either the forward/backward direction or in longer periods.
The nature of events to represent the same scene in various representations is particularly valuable in scenarios where only weak labels are available, since the regions of focus and confidence within the model can be distinct.
Considering that identifying reliable regions is a crucial part of WSSS where dense GT is not available, events offer the advantage of being able to represent various reliable regions for the same scene, depending on how they are represented in spatiotemporal dimension. 

Inspired by this aspect, we propose a framework based on dual-student learning.
To construct event stream pairs containing maximally dissimilar information while maintaining the same end-time, one stream is constructed to match the event representation ($E^{f}$) utilized during inference, while the other is composed by elongating the backward-obtained event stream to form $E^{b}$ ($\tau \ll \tau'$).

As depicted in Fig.~\ref{fig:overall_framework}, the proposed framework involves two branches of propagating the voxelized forward and backward event streams, $E^f$ and $E^b$, through distinct encoders $\phi^f$ and $\phi^b$, respectively. 
Subsequently, we define shallow U-Net~\cite{ronneberger2015u, Sun2022ESSLE}-like decoders $\mathcal{D}^f$ and $\mathcal{D}^b$ to acquire segmentation predictions $\mathbf{P}^f=\mathcal{D}^f(\mathbf{Z}^f)$ and $\mathbf{P}^b=\mathcal{D}^b(\mathbf{Z}^b)$, where both have the dimension of $\mathbb{R}^{C\times H \times W}$. 
Here, $H$ and $W$ are the height and width of the input event voxel, respectively.

The predictions are supervised with the given weak point labels $\mathbf{t}$ using pixel-wise cross-entropy loss. 
Given that most of the pixels are not annotated, we impose mere cross-entropy loss ($\text{CE}$) only on the pixels where the label exists.
This loss can be formally defined as follows:
\begin{equation} 
    \mathcal{L}_{weak} = -\frac{1}{2\vert\mathbf{t}\vert}\sum_{(x,y,c)\in\mathbf{t}} \{\text{CE}(\mathbf{P}^    f_{x,y},c) + \text{CE}(\mathbf{P}^b_{x,y},c)\}.
\end{equation}
where $(x,y,c)$ denotes the spatial position of the pixel and its class where weak point labels exist.
Accordingly, $\mathbf{P}^f_{x,y}\in\mathbb{R}^{C}$ represents the predicted logit of the forward branch at the position $(x,y)$, and $\mathbf{P}^b_{x,y}$ is similarly defined.
Note that we can use the same weak label as GT for both the forward and backward branches since the backward event voxel is reversed and the target time steps of the forward and backward events are the same.

However, as discussed above, the spatially sparse point labels are not sufficient to notify the concept of semantic segmentation for the model.
As a remedy, from the perspective of dual-student learning, we form a pseudo-GT based on the prediction of one branch and then use it as guidance for the opposite branch.
This helps two branches in our framework, forward and backward, learn from the temporally distinct data without direct feature fusion while building a consensus between individually optimizing models.
To ensure the quality of pseudo-GT, low-confidence prediction results should be rejected, especially at the early stage of training.
Formally, we first define reliability map $\mathbf{R}^f\in \mathbb{R}^{H \times W}$ as follows:
\begin{equation}\label{eq:reliability}
    \mathbf{R}^f_{x,y} = \text{max}(\mathbf{P}^f_{x,y}).
\end{equation}
Based on the reliability map, we generate pseudo-GT map $\mathbf{A}^f\in \mathbb{R}^{H \times W}$ by applying a predefined threshold $th$ to the reliability map as
\begin{equation}\label{eq:pgt}
    \mathbf{A}^f_{x,y} = 
\begin{cases}
    \argmax_{k}(\mathbf{P}^{f,k}_{x,y}), & \text{if } \mathbf{R}^f_{x,y}>th \\
    255 \quad (\text{ignoring index}),              & \text{else},
\end{cases}
\end{equation}
where $\mathbf{P}^{f,k}_{x,y}$ is the prediction score of $\mathbf{P}^f_{x,y}$ for $k^{th}$ class.
We omit the formation process of $\mathbf{R}^b$ and $\mathbf{A}^b$ for simplicity but undergo the same process.

With the obtained pseudo-GTs, we define dual loss as follows:
\begin{equation}
    \mathcal{L}_{dual} = \frac{1}{2}\text{CE}(\mathbf{P}^f,\mathbf{A}^b)+\frac{1}{2}\text{CE}(\mathbf{P}^b,\mathbf{A}^f).
\end{equation}
Note that the pseudo-GT from the forward branch guides the prediction of the backward branch, and vice versa.

\subsection{Feature-level Prototype-based Contrastive Learning}\label{sec:feat_proto}
In Section~\ref{method:dual}, we introduced a method to effectively utilize the unique strength of event data -- information inherently embedded in the temporal axis -- for semantic segmentation tasks through dual-student learning between original forward event data and backward event data. 
Despite the potential effectiveness of this data-centric approach, a challenge arises due to the inherent sparsity of point supervision in the proposed WSSS framework.
While reliable dense pseudo-GT is generated during the dual-student learning process to train the opposite branch, it primarily serves as supervision for the classification task at the logit level only.
Consequently, this approach falls short in guiding the models to learn which areas should be grouped into a single segment from a segmentation perspective.

\begin{figure*}[t]
    \centering\includegraphics[width=1.0\linewidth]{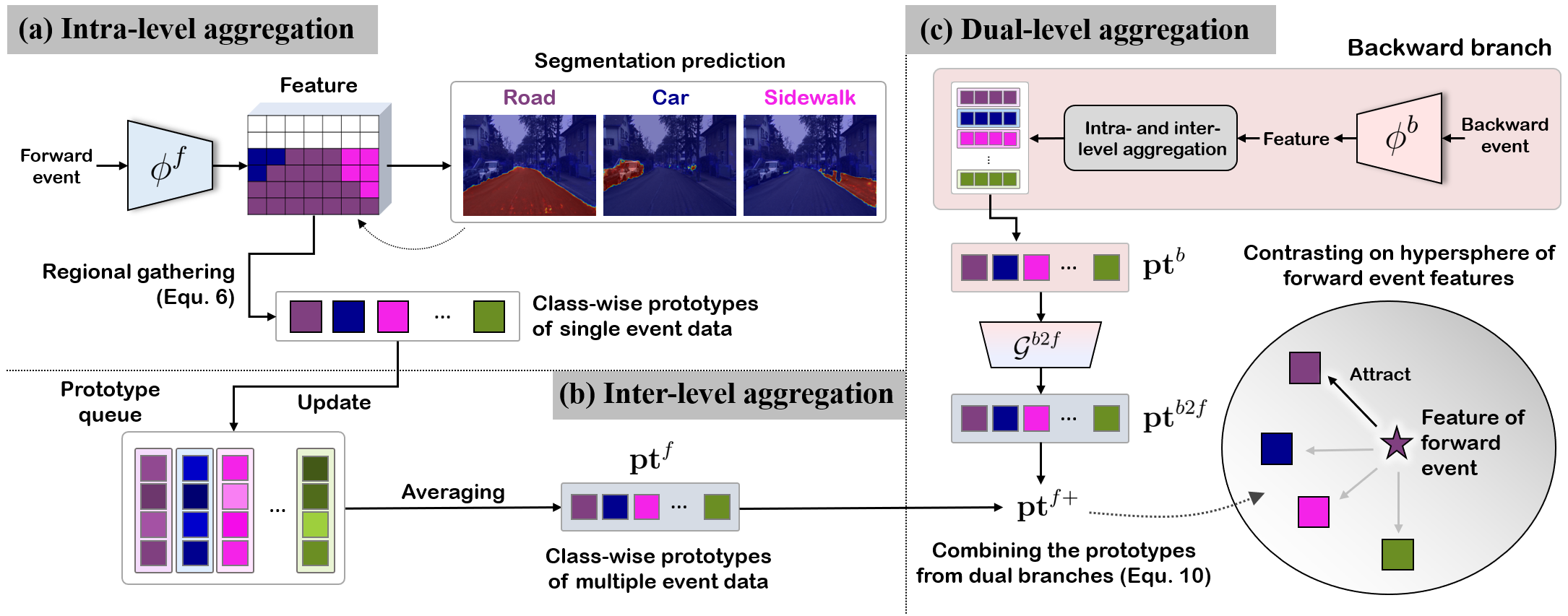}
    \caption{Visualization of the proposed prototype-based contrastive learning approaches based on the aggregations performed in three different levels.}
    \label{fig:proto}
\end{figure*}

To address this issue, we propose an approach based on feature-level contrastive learning, a technique actively utilized in segmentation learning. 
Given that event data is spatially more sparse and semantic perception is more challenging compared to RGB image data, we anticipate that feature-level contrastive learning will be highly effective.
Specifically, we adopt a strategy of defining class-wise prototypes for feature-level contrastive learning. 
We identify reliable regions for each class based on the predictions of the proposed model and aggregate features from those regions to obtain class-wise prototypes. 
These prototypes subsequently serve as anchors during contrasting features of each pixel.

In the proposed framework, to generate more representative prototypes, we conduct aggregation at both \textbf{intra-} and \textbf{inter-}levels, as depicted in Fig.~\ref{fig:proto}.
Throughout this section, for convenience, we will show the aggregation in the forward branch only, and the same process is performed in the backward branch.
Firstly, as an intra-level aggregation, we gather the features of the regions with a certain level of reliability, rejecting the features of the confusing regions.
We perform this process within a single event data during a specific iteration using $\mathbf{A}^f$, the reliable pseudo-GT of the forward branch according to the semantic segmentation predictions, as defined in Equ.~\ref{eq:pgt}.
Formally, the intra-level aggregation can be expressed as the following equation.
\begin{equation}
    \mathbf{pt}^{f,k} =  \sum \mathbf{R}^{f}_{x,y}\mathbf{Z}^{f}_{x,y} \quad \forall (x,y) \: \text{such that} \: \mathbf{A}^f_{x,y}=k,
\end{equation}
where $\mathbf{pt}^{f,k}$ is the forward branch's prototype of $k$-th class aggregated in intra-level and $\mathbf{Z}^{f}_{x,y}$ denotes the feature at pixel $(x,y)$.
Note that we utilize the reliability score $\mathbf{R}^{f}$ defined in Equ.~\ref{eq:reliability} for soft aggregation of the features.
For effective learning of the representation space, the features are normalized before and after aggregation to constrain them to lie on a hypersphere.

Compared to selecting only one point to acquire features, this region-based approach allows for the extraction of more robust class-wise prototypes, effectively representing each class in the event data used as input. 
However, if prototype-based contrasting is done within a single event data, there would be a high likelihood that features of the same class and class-wise prototypes are already close, possibly reducing the benefit of the contrastive approach.
Additionally, model predictions on event data may not be entirely reliable, especially for complex scenes early in the learning process.
Therefore, for more effective contrastive learning, we believe that it is necessary to acquire prototypes that represent the overall semantics rather than being dominated by one data.

To this end, we devised inter-level aggregation. 
Specifically, as training progresses, we stack the previously described intra-level aggregated class-wise prototypes in a queue. 
Then, we perform inter-level aggregation by averaging the stacked prototypes.
This approach enables us to utilize prototypes obtained from various event data encountered previously, acting as a memory-based approach for not only dynamic but also more stable representation learning.

The class-wise prototypes obtained through intra- and inter-level aggregation are utilized to contrast each feature. Leveraging the dual-student framework described in Section~\ref{method:dual} and using the pseudo-GT as a reference, each feature is encouraged to be close to the prototype corresponding to its class and distant from prototypes of other classes.
We formulate this process using the InfoNCE loss~\cite{oord2018representation} to drive effective contrastive learning as follows:
\begin{equation}\label{eq:contrast}
    \mathcal{L}_{proto}^f = -\sum_{x,y} \log \frac{\text{Sim}(\mathbf{Z}_{x,y}^{f},\mathbf{pt}^{f,\alpha})/\beta}
    {\sum^{C}_{k=1} \text{Sim}(\mathbf{Z}_{x,y}^{f},\mathbf{pt}^{f,k})/\beta} \:\text{where}\: \alpha=\mathbf{A}^b_{x,y}.
\end{equation}
Here, a similarity metric $\text{Sim}(\cdot)$ is defined as $sim(q_1,q_2) = e^{q_1 \cdot q_2}$ and $\beta$ is temperature.
Note that when we perform contrastive learning with the features and class-wise prototypes in the forward branch, the target classes for contrasting ($\alpha$) are set by the pseudo-GT obtained from the \underline{backward} branch.
If $\alpha=255$, we simply do not include that pixel for computing $\mathcal{L}_{proto}$.

\subsection{Dual-Level Prototype Aggregation via Distillation}\label{sec:align}
Section~\ref{sec:feat_proto} presents feature-level prototype-based contrastive learning to address the inherent sparsity of point supervision provided in EV-WSSS.
We observe that this approach can effectively enhance the semantic understanding of our framework.
However, the contrasting process is independently performed at each branch (\textit{i.e.}, forward or backward), which may not fully excavate the complementary potential of our dual approach.
To integrate these complementary strengths of the dual-student learning framework within the prototype-based contrastive learning, we propose prototype-level distillation performed across the dual branch.
Our main motivation is that the prototype aggregated in the forward branch contains helpful information from the perspective of the reverse branch, and vice versa.
However, the feature spaces of each branch, where contrastive learning occurs, are individually optimized.
Therefore, the prototypes of one branch cannot be directly delivered to the other branch.

To overcome this, we begin with feature-level distillation between two branches, as simplified in Fig.~\ref{fig:proto}(c).
Here, given that the input data of each branch are inherently different, the features obtained by each branch should also be distinctive, which is the main philosophy of our dual approach.
Therefore, unlike conventional distillation approaches that aim to directly align the feature spaces of two models, our goal is to build a projection bridge between two feature spaces, preserving their unique characteristics while facilitating knowledge transfer.
Specifically, we define linear projection layers for each direction, \textit{i.e.}, forward-to-backward ($\mathcal{G}^{f2b}$) and backward-to-forward ($\mathcal{G}^{b2f}$).
The layers are optimized by the following feature-level distillation loss:
\begin{equation}\label{eq:distill}
    \mathcal{L}_{distill} = \frac{1}{2}\vert\mathcal{G}^{b2f}(\mathbf{Z}^b)-\mathbf{Z}^f\vert_1 + \frac{1}{2}\vert\mathcal{G}^{f2b}(\mathbf{Z}^f)-\mathbf{Z}^b\vert_1,
\end{equation}
where $\vert\cdot\vert$ denotes L1 loss.
Then, the projection layers function as a sort of translator between two individual feature spaces, transferring the helpful information from the perspective of each branch.
With these translators, we deliver the prototypes aggregated in one branch to the opposite branch as follows:

\begin{equation}
    \mathbf{pt}^{f2b} = \mathcal{G}^{f2b}(\mathbf{pt}^{f}) \quad \text{and} \quad \mathbf{pt}^{b2f} = \mathcal{G}^{b2f}(\mathbf{pt}^{b}).
\end{equation}

Finally, within each branch, we combine the original and delivered prototypes to form the ultimate class-wise prototypes for contrastive learning.
Formally,
\begin{equation}
\label{eq:pt+}
    \mathbf{pt}^{f+} = \mathbf{pt}^{f}+\mathbf{pt}^{b2f}.
\end{equation}
Although we omit, $\mathbf{pt}^{b+}$ is acquired in the same manner.
Subsequently, in Equ.~\ref{eq:contrast}, the obtained $\mathbf{pt}^{f+}$ and $\mathbf{pt}^{b+}$ including dual information substitutes $\mathbf{pt}^{f}$ and $\mathbf{pt}^{b}$, which was aggregated through single branch only.

To sum up, we perform three levels of aggregations in total.
First, intra-level aggregation aims to carefully extract regional information from a single event stream data, rejecting unreliable regions.
Second, inter-level aggregation is performed using memory-based approaches to consider the variances among multiple event streams.
Finally, dual-level aggregation across the branches leverages the benefit of dual-student learning via the feature-level projection of the prototypes, leading to even more effective representation.

\section{Experiments}
\subsection{Datasets}
We conduct our research using three datasets: DDD17-Seg~\cite{Alonso2018EVSegNetSS}, comprising 40 sequences with 15,950 training and 3,890 testing events, with a $346 \times 200$ resolution. DSEC-Semantic~\cite{Sun2022ESSLE} offers semantic annotations for 11 sequences from the DSEC~\cite{Gehrig2021DSECAS} dataset, with the training and testing splits including 8,082 and 2,809 events, of spatial size $640 \times 440$.
Additionally, to demonstrate the practicality of EV-WSSS, we validated the performance of the proposed framework using the DSEC Night-Semantic Dataset~\cite{Xia2023CMDACD}. 
However, this dataset provides dense annotation only on the test set for evaluation purposes. Therefore, we manually annotate the train set with our 1C1C setting, the DSEC Night-Point dataset. Following the DSEC-Semantic~\cite{Sun2022ESSLE}, we annotate the sparse point labels on events with 11 classes. 
In addition, we propose a 1C10C setting, which annotates 10 points per class.
Although this would require ten times the annotation burden, we find that this setting of EV-WSSS can achieve a performance comparable to fully supervised methods.
Details about DSEC N-P can be found in supp.

\subsection{Implementation Details}
We used an NVIDIA RTX 3090 for training and optimized the network through the RAdam~\cite{liu2019variance} optimizer. The threshold, $th$, used in Equ.~\ref{eq:pgt} is set to 0.5, and the temperature, $\beta$, used in Equ.~\ref{eq:contrast} is set to 0.1. The criterion for stacking reverse events, \(\tau'\), is set to five times the number of events compared to when stacking with \(\tau\). 
Additional analysis on these hyperparameters can be found in supp.

\subsection{Ablation Studies}

In this paper, we introduce two main approaches: dual-student learning (Section \ref{method:dual}) and feature-level contrastive learning based on class-wise prototypes (Section~\ref{sec:feat_proto}-\ref{sec:align}).
To demonstrate the effectiveness of the proposed approaches, we conduct ablation studies about each component.
The experiment is performed on DSEC-Semantic~\cite{Sun2022ESSLE} under an event-only setting.
We set our baseline as a model with a single forward branch, trained by sparse point supervision only.

\begin{table}[t]
\centering
\caption{Ablation study on the proposed modules. Here, $\mathcal{L}_{proto}^{+}$ represents the scenario where $pt^{f+}$ and $pt^{b+}$ from Equ.~\ref{eq:pt+} are employed for contrastive learning in Equ.~\ref{eq:contrast}.}
\setlength\tabcolsep{6.5pt}
\label{tb:abl_component_analysis}
\resizebox{0.55\linewidth}{!}{
\begin{tabular}{ccccc|c} 
\hline
\rowcolor[rgb]{0.8,0.8,0.8} & \multicolumn{1}{c}{$\mathcal{L}_{dual}$} & \multicolumn{1}{c}{$\mathcal{L}_{proto}$} & $\mathcal{L}_{distill}$ & $\mathcal{L}_{proto}^{+}$ & mIoU (\%)    \\ \hline
(a)& &  & & &  39.1 \\ \hline
(b)& \checkmark &  & & &  42.6 \\ \hline
(c)& \checkmark & \checkmark &  & &  43.6  \\ \hline
(d)& \checkmark & \checkmark & \checkmark &       &44.7 \\ \hline
(e)& \checkmark & \checkmark & \checkmark &  \checkmark    &  45.6 \\
\hline
\end{tabular}
}
\end{table}

\begin{figure*}[t]    \centering\includegraphics[width=0.8\linewidth]{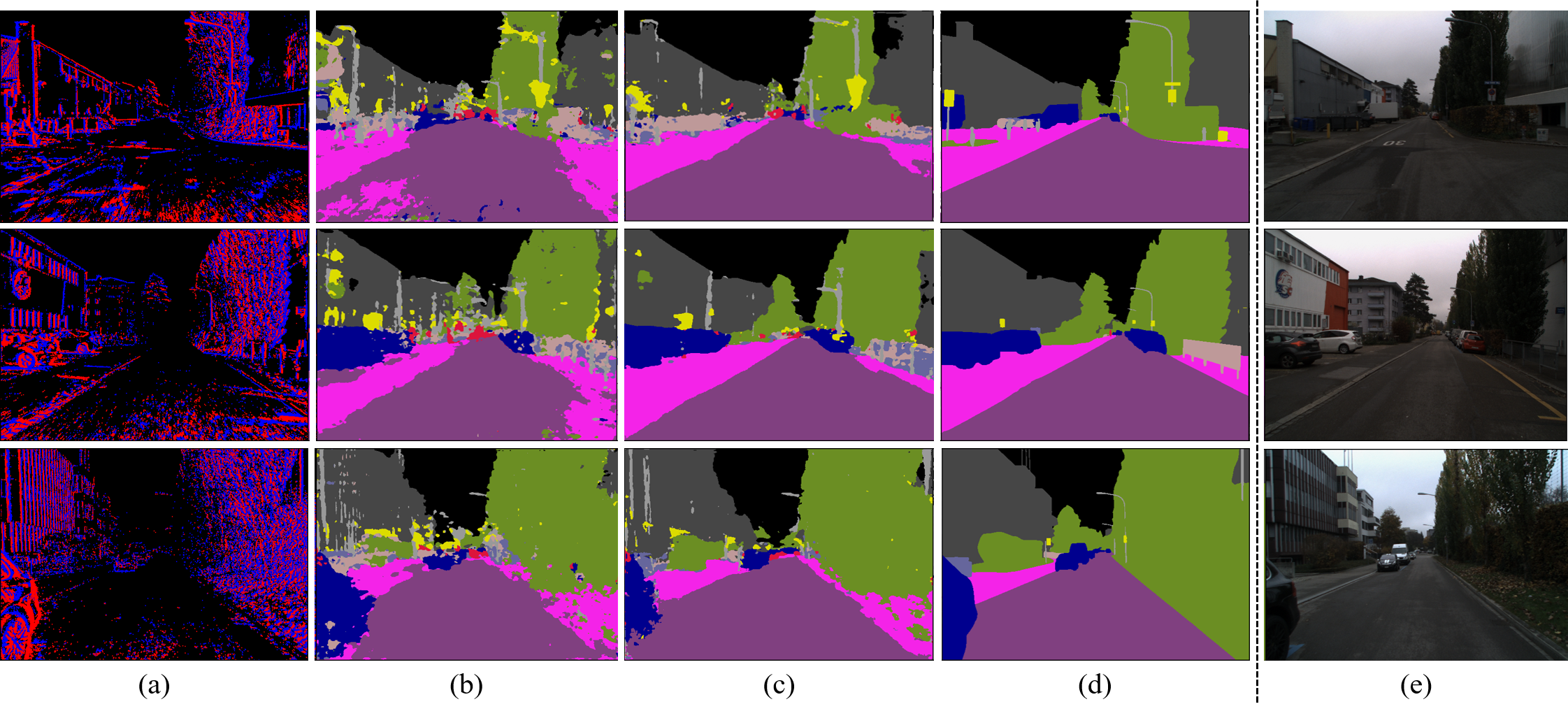}
    \caption{Qualitative ablation of EV-WSSS framework. (a) visualized event data, (b) results of baseline, (c) results of our final model, (d) segmentation GT, and (e) image.}
    \label{fig:qual_dsec}
\end{figure*}

As in Table.~\ref{tb:abl_component_analysis}, we verify our dual-student learning framework (b) outperforms the baseline (a) by a large margin.
It shows that our approach effectively leverages the future information of event data during training.
Moreover, the experiment (c) with the proposed prototype-based distillation approach brings meaningful performance gain.
It supports that our intra- and inter-level aggregation leads to the robust perception of semantic representation, even under extremely challenging conditions with the sparseness of both event data and weak labels.
Finally, (e), with dual-level prototype aggregation via distillation, the model achieves even better performance.
A comparison between (d) and (e) verifies that (e)'s performance gain comes not only from the distillation between the two branches but also from the dual-level aggregation during class-wise prototype generation.
We provide qualitative ablation in Fig.~\ref{fig:qual_dsec}.

\begin{figure}[t]
  \centering
\begin{subfigure}[b]{0.4\textwidth}
    \centering
    \resizebox{.9\linewidth}{!}{
        \begin{tabular}{l|c}
        \hline
        \rowcolor[rgb]{0.8,0.8,0.8} Method            & mIoU \\ \hline
        (I) Baseline          &  39.1    \\ 
        (II) Self-supervised   &   40.4   \\ 
        (III) Self-supervised (EMA~\cite{he2020momentum}) &  41.0    \\ \hline
        (IV) Dual (Ours)       &  42.6  \\ \hline
        \end{tabular}}
    \caption{}
    \label{tb:selfsup_dual}
  \end{subfigure}\hfill
\begin{subfigure}[b]{0.6\textwidth}
    \centering
    \includegraphics[width=\textwidth]{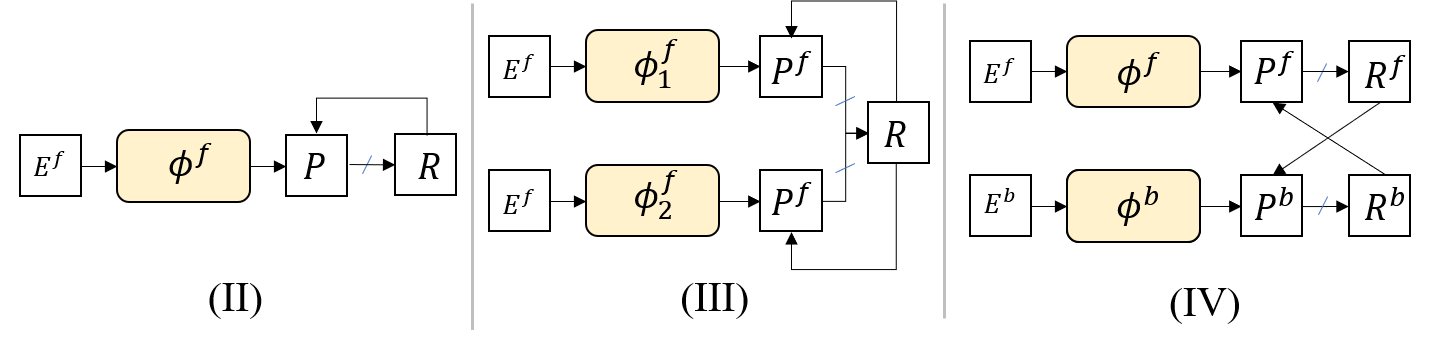}
    \caption{}
    \label{fig:selfsup_dual}
  \end{subfigure}
  \caption{(a) Comparison with various self-supervised approaches and (b) the simplified training pipeline for the respective approaches. For (IV), we provide the performance of ours without the prototype-related components for a fair comparison.}
  \label{fig:compare_selfsup_dual}
\end{figure}

To more comprehensively analyze the dual-student learning framework from the methodology perspective, we conduct a comparative experiment comparing our method with other possible options for pseudo-GT generation during training. 
Specifically, as shown in the right diagrams of Fig.~\ref{tb:selfsup_dual},  we test two simple yet effective methods: (II) direct self-supervision and (III) guidance from an EMA-based teacher~\cite{he2020momentum}.
In (II), the segmentation prediction of the forward branch is processed into pseudo-GT, similar to the Equ.~\ref{eq:pgt}.
On the other hand, (III) defines a teacher network updated by exponential moving average (EMA) from the main network only, where the prediction of the teacher network directly guides the main network.
Finally, the proposed approach (IV) defines a dual branch of the forward and backward event data, which mutually guide each other.
The table at the left of Fig.~\ref{tb:selfsup_dual} supports that our dual formulation excavates the benefit of the event data, leveraging its temporal aspects effectively.

\subsection{Comparisons with the Other Methods}
\label{sec:experiments}
\textbf{DSEC-Semantic and DDD17 Datasets.}
Table~\ref{tab:quan_dsec} and Table~\ref{tab:quan_ddd} demonstrate comparisons on the DSEC and DDD17 datasets among fully supervised, UDA, and our weakly supervised setting. 
We provide the performance of the fully supervised methods~\cite{Alonso2018EVSegNetSS, Wang2021DualTL, Wang2021PyramidVT, Hamaguchi2023HierarchicalNM, Biswas2022HALSIEHA, Kim2021BeyondCD, jia2023event} using dense GT as reference for the current state of event-based segmentation.
Meanwhile, UDA methods~\cite{Sun2022ESSLE, Wang2021EvDistillAE, Rebecq2019HighSA, Messikommer2021BridgingTG, Gehrig2019VideoTE} are similar to ours in that they do not require full GT in the target domain; however, they assume the existence of the source dataset for pre-training, which is less practical in our view. 
Due to these differences, we are aware that EV-WSSS cannot be directly comparable with the methods.
Nevertheless, our approach's substantial performance supports its potential as a promising option for event-based segmentation, especially in a practical event-only setting.
Notably, on the DDD17 dataset, EV-WSSS with 1C10C setting delivers promising results, surpassing all UDA performances with just weak labels, and is even ahead of or comparable to some supervision-based methods~\cite{Alonso2018EVSegNetSS, Wang2021DualTL, Wang2021PyramidVT, Kim2021BeyondCD, jia2023event}.

\begin{table}[t]
    \caption{Comparison with other event-based semantic segmentation methods under the fully-supervised and unsupervised domain adaptation methods on DSEC dataset.}
    \setlength\tabcolsep{10.5pt}
    \begin{minipage}[b]{.47\linewidth}
        \centering
        \resizebox{0.99\linewidth}{!}{
            \begin{tabular}{ccc}
            \hline
            \rowcolor[rgb]{0.75,0.75,0.75}Method   & Publication & mIoU \\ \hline
            \rowcolor[rgb]{0.95,0.95,0.95}
            \multicolumn{3}{l}{Fully Supervised} \\ \hline
            Ev-SegNet~\cite{Alonso2018EVSegNetSS} & CVPRW'19 & 51.76 \\
            ESS-sup~\cite{Sun2022ESSLE} & ECCV'22 & 53.29\\
            HMNet-B~\cite{Hamaguchi2023HierarchicalNM} & CVPR'23 & 51.20 \\
            HMNet-L~\cite{Hamaguchi2023HierarchicalNM} & CVPR'23 & 55.00 \\
            HALSIE~\cite{Biswas2022HALSIEHA} & WACV'24  & 52.43 
              \\\hline
            \end{tabular}
        }
    \end{minipage}
    \hfill
    \begin{minipage}[b]{.47\linewidth}
        \centering
        \resizebox{0.99\linewidth}{!}{
            \begin{tabular}{ccc}
            \hline
            \rowcolor[rgb]{0.75,0.75,0.75}Method   & Publication & mIoU \\ \hline
            \rowcolor[rgb]{0.95,0.95,0.95}
            \multicolumn{3}{l}{Unsupervised Domain Adaptation} \\ \hline
            E2VID~\cite{Rebecq2019HighSA} & TPAMI'19 & 44.08 \\
            EV-Transfer~\cite{Messikommer2021BridgingTG} & RA-L'22 & 24.37\\
            ESS~\cite{Sun2022ESSLE} & ECCV'22 & 45.38 \\
            \hline
            \rowcolor[rgb]{0.95,0.95,0.95}
            \multicolumn{3}{l}{Weakly Supervised} \\ \hline
            EV-WSSS (1C1C) & Ours  & 45.55  
              \\
            EV-WSSS (1C10C) & Ours  & 51.19  
              \\
              \hline
            \end{tabular}
        }
    \end{minipage}
    \label{tab:quan_dsec}
\end{table}

\begin{table}[t]
    \caption{Comparison with other event-based semantic segmentation methods under the fully-supervised and unsupervised domain adaptation methods on DDD17 dataset.}
    \setlength\tabcolsep{6.5pt}
    \centering
    \begin{minipage}[b]{.46\linewidth}
        \centering
        \resizebox{0.96\linewidth}{!}{
            \begin{tabular}{ccc}
            \hline
            \rowcolor[rgb]{0.75,0.75,0.75}Method   & Publication & mIoU \\ \hline
            \rowcolor[rgb]{0.95,0.95,0.95}
            \multicolumn{3}{l}{Fully supervised} \\ \hline
            Ev-SegNet~\cite{Alonso2018EVSegNetSS} & CVPRW'19 & 54.81 \\
            DTL~\cite{Wang2021DualTL} & ICCV'21 & 58.80 \\
            PVT-FPN~\cite{Wang2021PyramidVT} & ICCV’21 & 53.89 \\
            SpikingFCN~\cite{Kim2021BeyondCD} & NCE’22  & 34.20 \\
            ESS-sup~\cite{Sun2022ESSLE} & ECCV'22  & 61.37\\
            EvSegformer~\cite{jia2023event} & TIP'23  & 54.41 \\
            HALSIE~\cite{Biswas2022HALSIEHA} & WACV'24 & 60.66 
              \\\hline
            \end{tabular}
        }
    \end{minipage}
    \hfill
    \begin{minipage}[b]{.48\linewidth}
        \centering
        \resizebox{0.95\linewidth}{!}{
            \begin{tabular}{ccc}
            \hline
            \rowcolor[rgb]{0.75,0.75,0.75}Method   & Publication &  mIoU \\ \hline
            \rowcolor[rgb]{0.95,0.95,0.95}
            \multicolumn{3}{l}{Unsupervised Domain Adaptation} \\ \hline
            E2VID~\cite{Rebecq2019HighSA} & TPAMI'19 & 48.47 \\
            Vid2E~\cite{Gehrig2019VideoTE} & CVPR'20 & 45.48 \\
            EV-Transfer~\cite{Messikommer2021BridgingTG} & RA-L'22 & 15.52 \\
            ESS~\cite{Sun2022ESSLE} & ECCV'22  & 53.09 \\
            \hline
            \rowcolor[rgb]{0.95,0.95,0.95}
            \multicolumn{3}{l}{Weakly supervised} \\ \hline
            EV-WSSS (1C1C) & Ours & 56.88
              \\
            EV-WSSS (1C10C) & Ours & 60.39
              \\ \hline
            \end{tabular}
        }
    \end{minipage}
    \label{tab:quan_ddd}
\end{table}

\begin{table}[t]
\centering
\caption{Comparisons on DSEC N-P between the results of ESS~\cite{Sun2022ESSLE}, our baseline, ours with dual loss only, and our final model. We train ESS on DSEC N-S dataset.}
\label{tb:dsec-night}
\resizebox{0.99\linewidth}{!}{%
\renewcommand{\tabcolsep}{3pt}
\begin{tabular}{c|ccccccccccc|c}
\hline
\rowcolor[rgb]{0.75,0.75,0.75}
Method           & Sky & Building & Fence & Person & Pole & Road & Sidewalk & Vegetation & Car & Wall & Traffic-sign & mIoU \\ \hline
ESS*~\cite{Sun2022ESSLE}            & 6.1 & 22.1 & 0.0  & 3.6  & 3.5  & 67.8  & 13.5  & 19.7  & 16.6  & 0.0   & 1.6   & 14.0    \\
Baseline ($\mathcal{L}_{weak}$ only)  & 62.5  & 40.3  & 2.4  & 6.8  & 10.4  & 78.8  & 29.8  & 40.9  & 30.9  & 9.3   & 10.5   & 29.4     \\
Ours ($\mathcal{L}_{weak}+\mathcal{L}_{dual}$) & 70.7  & 42.5  & 0.8  & 8.5  & 15.2  & 84.5  & 37.5  & 46.5  & 33.5  & 9.7   & 12.1   & 32.9     \\
\hline
Ours             & 73.8  &  41.1 & 2.7  & 13.2 & 19.0  & 86.2  & 39.4  & 51.8  & 39.1  & 19.7   & 14.2   & 36.4     \\ \hline
\end{tabular}}
\label{tab:quan_night_point}
\end{table}

\begin{figure*}[t]
    \centering\includegraphics[width=0.9\linewidth]{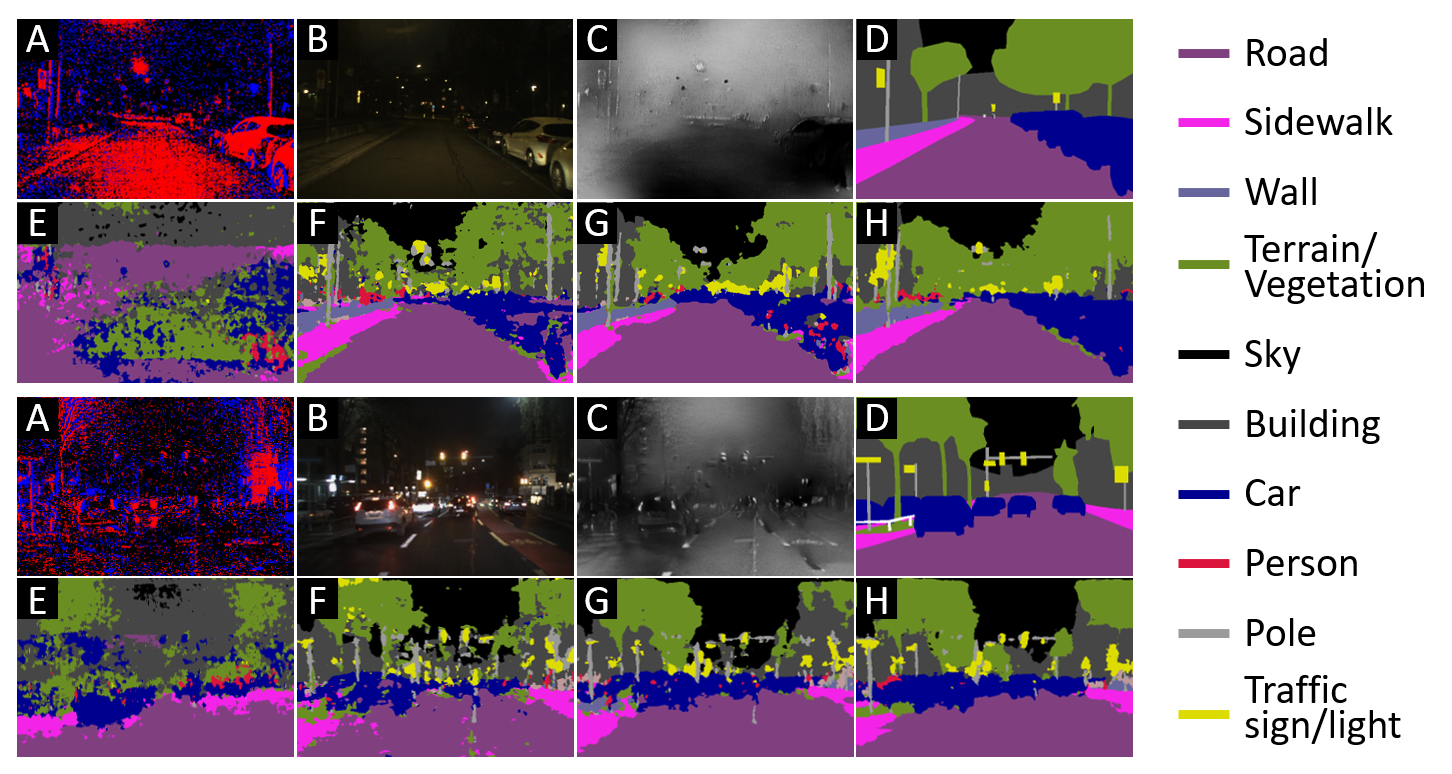}
    \caption{Qualitative comparisons on DSEC Night-Point dataset. From A to D: visualized event data, image (for reference only), image reconstructed from event by ESS~\cite{Sun2022ESSLE}, and GT semantic labels. From E to H: results of ESS~\cite{Sun2022ESSLE}, baseline with weak supervision only, baseline with dual-student learning, and our final framework.}
    \label{fig:qual_night}
\end{figure*}

\noindent
\textbf{DSEC Night-Point Dataset.}
We explore the benefits of EV-WSSS by training it on the DSEC Night-Point Dataset, an edge case with nighttime events, and present the outcomes in Table~\ref{tab:quan_night_point}.
Since ESS~\cite{Sun2022ESSLE} falls under a UDA setting that allows for training without GT in the target domain, we document its performance alongside ours. ESS, which utilizes a pre-trained E2VID~\cite{Rebecq2019HighSA} network, often fails to train properly due to artifacts arising in domains (\ie,~night noisy events) significantly different from its training data. 
Our event-only WSSS setting, despite the noisiness of night scene events, achieves promising performance. 
This difference can be more clearly observed qualitatively in Fig.~\ref{fig:qual_night}.


\section{Conclusion}
This paper introduces EV-WSSS: a pioneering weakly supervised methodology for event-based semantic segmentation leveraging sparse point annotations.
We present asymmetric dual-student learning, exploiting both forward and reversed event data to enjoy the rich temporal information inherent in events, capturing complementary knowledge from different time perspectives. 
To overcome the obstacles presented by sparse supervision, our framework incorporates feature-level contrastive learning, employing class-wise prototypes aggregated meticulously across three levels: intra, inter, and dual.
We verify our framework across several benchmarks, including the newly introduced DSEC Night-Point dataset with sparse point annotations.
The results show that EV-WSSS achieves remarkable performance, significantly reducing the dependence on dense annotations. 

%
%

\noindent\textbf{Acknowledgement}

\noindent This work was supported by the National Research Foundation of Korea(NRF) grant funded by the Korea government(MSIT) (NRF2022R1A2B5B03002636).

\bibliographystyle{splncs04}
\bibliography{main}

\title{Finding Meaning in Points: Weakly Supervised Semantic Segmentation for Event Cameras: \textit{Supplementary Materials}} 

\titlerunning{Finding Meaning in Points: WSSS for Event Cameras}


\author{Hoonhee Cho\orcidlink{0000-0003-0896-6793}\thanks{Equal contribution.}, Sung-Hoon Yoon\orcidlink{0000-0001-5851-2031}\protect\CoAuthorMark, Hyeokjun Kweon\orcidlink{0000-0003-4442-5513}\protect\CoAuthorMark, \\
and Kuk-Jin Yoon\orcidlink{0000-0002-1634-2756}}

\authorrunning{Cho et al.}

\institute{Visual Intelligence Lab., KAIST \\
\email{\{gnsgnsgml, yoon307, 0327june, kjyoon\}@kaist.ac.kr}
}

\maketitle
\renewcommand{\thefigure}{A\arabic{figure}}
\renewcommand{\thetable}{A\arabic{table}}
\renewcommand\thesection{\Alph{section}}

\newcolumntype{P}[1]{>{\centering\arraybackslash}p{#1}}

\section{Details about DSEC Night-Point}~\label{sec:protocol}

\begin{figure}[h]
    \centering
    \includegraphics[width=0.85\linewidth]{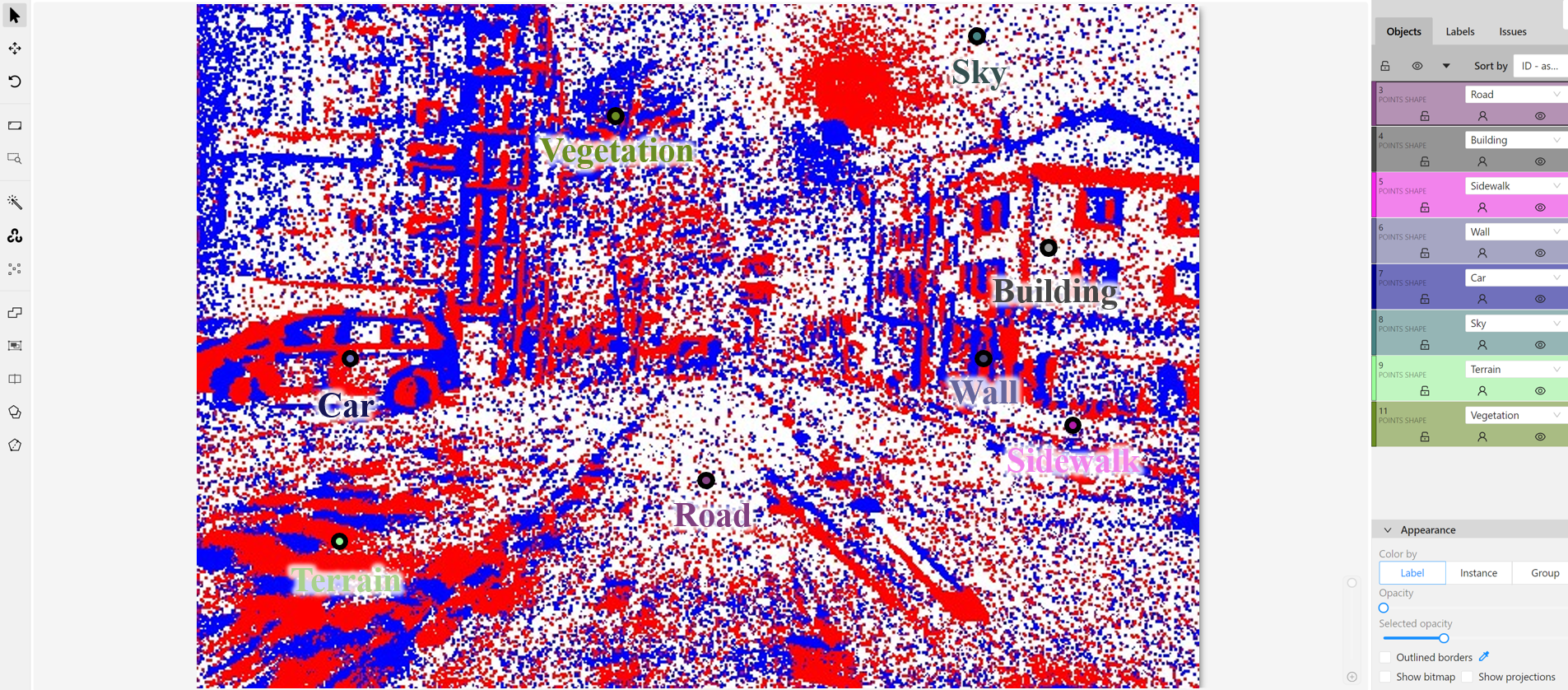}
    \caption{Demonstration of point annotation process on DSEC-Night dataset.}
    \label{fig:label_cvat}
\end{figure}

The proposed Event-based Weakly Supervised Semantic Segmentation (EV-WSSS) method performs semantic segmentation with only one point-level supervision per existing class (referred to as "1-Class-1-Click" or 1C1C) for the given event voxel.
We demonstrated the effectiveness of EV-WSSS through the DSEC-Semantic~\cite{Sun2022ESSLE} dataset, which is a widely used dataset in event-based semantic segmentation. 
Furthermore, to emphasize the strength of weakly supervised approaches in situations where dense labeling using images is challenging, we validated our EV-WSSS in nighttime environments.
For this, we constructed a novel dataset named DSEC Night-Point, composed of raw event data from DSEC-Night and weak labels newly annotated under our "1C1C" setting. 

The detailed protocol for acquiring the DSEC Night-Point is as follows.
Since the original DSEC-Night~\cite{Xia2023CMDACD, Gehrig2021DSECAS} dataset is densely annotated only for the \textit{validation} set, we annotate point labels for the \textit{train} set. 
The \textit{train} set is divided into five splits according to the DSEC-Night~\cite{Xia2023CMDACD, Gehrig2021DSECAS} as `Zurich City 09a', `Zurich City 09b', `Zurich City 09c', `Zurich City 09d', and `Zurich City 09e'.
Each split respectively comprises 508, 109, 371, 478, and 226 frames, totaling 1,692 frames.

For the annotation process, we employ five computer vision experts, one per split.
Before conducting the annotation, each annotator reviewed the whole visualized event stream, just like watching a video.
This enables the annotators to acquire prior knowledge, \textit{e.g.}, overall scene outline or objects existing in the scene, about the given split.
Subsequently, the annotators are requested to annotate each event frame with the 1C1C setting.
For this, we utilized an online tool (CVAT~\cite{sekachev2019computer}) as shown in~\cref{fig:label_cvat}.
To ensure labeling consistency between annotators in controversial regions, we incorporated discussions among annotators during the labeling process to reach a consensus.
Further, as the accuracy of the point annotation is crucial, we request the annotators to ignore the confusing cases rather than provide a possibly incorrect label.
After completing the labeling process, each annotator conducted cross-validation on the other splits.

Note that, throughout the annotation process, only the visualized event streams are exclusively used, unlike the labeling process~\cite{Alonso2018EVSegNetSS} where paired images were used. 
However, our annotation process can further utilize images, if it possibly enables more accurate labeling. 

We also investigate the burden of our annotation process.
In our scenario focused solely on events, obtaining precise labeling can pose challenges, primarily because accurately discerning object boundaries can be difficult. 
Conversely, our weakly supervised approach with the 1C1C setting necessitates only a single point annotation for each class, involving a straightforward marking within the objects' interiors.
Quantitatively, when labeling the DSEC-Night dataset under the 1C1C condition, an average of \textbf{50 seconds} per frame is required. 
Considering that the annotation time for driving scenes with dense pixel-level labels requires more than 1.5 hours~\cite{cordts2016cityscapes} and more than 2 hours in the DSEC-Night dataset, it underscores the practicality and importance of weak labels in the context of event-based semantic segmentation.

\renewcommand{\thefigure}{B\arabic{figure}}
\renewcommand{\thetable}{B\arabic{table}}

\section{Hyperparameters Analysis}

\begin{table}[h]
\centering
\caption{Results according to different threshold ($th$).}
\setlength\tabcolsep{7.2pt}
\label{tb:abl_threshold}
\resizebox{0.62\linewidth}{!}{
\begin{tabular}{lccccc}
\hline
\rowcolor[rgb]{0.8,0.8,0.8} $th$   & 0.3   & 0.4  & 0.5   & 0.6   & 0.7   \\ \hline
mIoU & 45.32 & 45.2 & 45.55 & 44.83 & 44.73 \\ \hline
\end{tabular}
}
\end{table}

The proposed method involves two notable hyperparameters: the threshold ($th$) for dual-student learning in Eq.~\textcolor{red}{4} and the temperature ($\beta$) for prototype-based contrastive learning in Eq.~\textcolor{red}{7} of the main paper.
To check the impact of the change in these parameters on the performance of our framework, we conduct experiments while adjusting them.
First, Table~\ref{tb:abl_threshold} shows the performance of the EV-WSSS framework with various threshold values.
The results implied that the proposed method is robust to the threshold for the dual-student learning approach, underscoring the effectiveness of our framework.
Further, we tested various temperature values (1.0 and 0.5, where 0.1 is our default).
We confirm that the change in temperature does not have much effect on the final performance ($\pm$1\% in mIoU).

%

\renewcommand{\thefigure}{C\arabic{figure}}
\renewcommand{\thetable}{C\arabic{table}}

\section{Experiments on Incomplete and Noisy Annotation}
Unlike the fully-supervised semantic segmentation approaches that can benefit from dense pixel-level GTs, our EV-WSSS only access weak supervision.
Therefore, the accuracy of each point label is crucial, similar to the other pointly-supervised approaches in the imagery domain~\cite{cheng2022pointly,li2023point2mask,fan2022pointly}.
To verify that our method can perform robustly against the errors innated in weak labels, we conduct an experiment by modeling the possible source of noises involved during the annotation process. We model the noise as (I) absence of annotation and (II) wrongly annotated class.

\subsection{Case (I): Incomplete Annotation}
\begin{table}[h]
    \caption{Experimental result on 1C1C setting with incomplete annotation using DSEC datasets. $\mathcal{W}^{target}$ denotes the weak point-level GT in the target domain. `Incomplete' indicates that a 10\% drop rate was applied to the confusing classes~(\textit{wall}, \textit{fence}, \textit{person}, and \textit{traffic sign}).}
    \renewcommand{\arraystretch}{1.05}
    \setlength\tabcolsep{6.5pt}
    \centering
    \resizebox{0.81\linewidth}{!}{
    \begin{tabular}{p{0.15\linewidth} P{0.35\linewidth} |P{0.2\linewidth}}  
    \hline
     \rowcolor[rgb]{0.75,0.75,0.75}
    Method & Used GT Type & mIoU 
    \\
    \hline
    \rowcolor[rgb]{0.95,0.95,0.95}
    \multicolumn{3}{c}{Weakly-supervised (1C1C)} \\ \hline
    EV-WSSS &  $\mathbf{\mathcal{W}^{target}}$  & 45.55 \\
    EV-WSSS & Incomplete $\mathbf{\mathcal{W}^{target}}$ & 44.10 (-1.4) 
    \\
    \hline
    \end{tabular}
    }
    \label{tab:noisy_labels}
\end{table}

As mentioned in Section~\ref{sec:protocol}, our annotation protocol for the 1C1C setting requests that the annotator ignore the confusing cases rather than provide a possibly incorrect label.
Here, the degree of confusion can differ across different categories.
For example, the frequent and large classes (\textit{road}, \textit{car}, etc.) are much easier to label than the confusing classes (\textit{wall}, \textit{fence}, \textit{person}, and \textit{traffic sign}).
Considering the above, we model the incomplete annotation by discarding some point labels of the confusing classes from each event frame, with a drop rate of 10\%. 
The results shown in Table~\ref{tab:noisy_labels} demonstrate that our EV-WSSS still achieves substantial segmentation performance, even with incomplete annotation.
This highlights the practicality of the proposed weakly supervised approach.

\subsection{Case (II): Wrong Annotation}
\begin{table}[h]
    \caption{
    Experimental result on 1C1C setting with noisy annotation using DSEC.}
    \setlength\tabcolsep{15.5pt}
    \centering
        \resizebox{0.7\linewidth}{!}{
            \begin{tabular}{c|ccc}
            \hline
            \rowcolor[rgb]{0.95,0.95,0.95} $p$ & 0 (ours) & 0.1 & 0.2
            \\
            \hline
            mIoU & 45.55 & 44.73 &  44.33  
            \\
            \hline
            \end{tabular}
        }
    \label{tab:noise}
\end{table}

Table~\ref{tab:noise} provides the results of additional experiments regarding case (II). 
To model the confusion between classes that possibly occurs during the annotation process, We randomly chose two point labels from each event stream and swapped their classes with a probability of $p$. 
Although performance declines as $p$ increases, the reduction is not severe, suggesting the robustness of EV-WSSS against the annotation noise.

\renewcommand{\thefigure}{D\arabic{figure}}
\renewcommand{\thetable}{D\arabic{table}}

\section{Unsupervised Domain Adaptation (UDA) using Weak Labels in the Target Domain}
Table~\ref{tab:ess_wsss} provides the results of ESS~\cite{Sun2022ESSLE} with weak labels.
On DSEC, which has a smaller domain gap with ESS's source, the additional use of weak labels indeed degrades performance compared to ESS alone. 
This implies that the naive use of weak labels may not provide a meaningful benefit over the gain from the source GT, highlighting the necessity of our approach even from the perspective of UDA.
Meanwhile, on DSEC N-P, ESS itself struggles with a severe domain gap. 
Although the additional use of weak labels mitigates this to some extent, the achieved performance is still far lower than the weak baseline. 
Conversely, EV-WSSS consistently outperforms all of them.
To sum up, all the results clearly demonstrate that our contribution lies in fully utilizing weak labels, rather than simply relying on them.

\begin{table}[h]
    \vspace{-15pt}
    \caption{Comparison with weakly supervised UDA methods.}
    \setlength\tabcolsep{6.5pt}
    \centering
        \resizebox{0.95\linewidth}{!}{
            \begin{tabular}{l|cccc}
            \hline
            \rowcolor[rgb]{0.95,0.95,0.95} Backbone  & ESS~\cite{Sun2022ESSLE} & ESS~\cite{Sun2022ESSLE} + Weak Sup. & Weak Sup & Ours \\ \hline
            DSEC & 45.38 &  42.45 & 39.06 &45.55 
            \\
            DSEC N-P & 13.97 & 26.38 &  29.35 & 36.40
            \\
            \hline
            \end{tabular}
        }
    \label{tab:ess_wsss}
    \vspace{-15pt}
\end{table}

\renewcommand{\thefigure}{E\arabic{figure}}
\renewcommand{\thetable}{E\arabic{table}}



\clearpage

\end{document}